\documentclass[11pt,a4paper]{article}
\usepackage[hyperref]{acl2020}
\usepackage{times}
\usepackage{latexsym}
\usepackage{multirow}
\usepackage{multicol}
\usepackage{graphicx}

\aclfinalcopy
\usepackage{microtype}
\usepackage[colorinlistoftodos,prependcaption,textsize=tiny]{todonotes}
\def\aclpaperid{1808} 
\author{Sinong Wang, Madian Khabsa, Hao Ma \\
  Faceboook AI, Seattle, USA\\
  \texttt{\{sinongwang, mkhabsa, haom\}@fb.com}}

\title{To Pretrain or Not to Pretrain: Examining the Benefits of Pretraining on Resource Rich Tasks}

\date{}

\begin{document}
\maketitle
\begin{abstract}
Pretraining NLP models with variants of Masked Language Model (MLM) objectives  has recently led to a significant improvements on many tasks. This paper examines the benefits of pretrained models as a function of the number of training samples used in the downstream task. On several text classification tasks, we show that as the number of training examples grow into the millions, the accuracy gap between finetuning BERT-based model and training vanilla LSTM from scratch narrows to within 1\%. Our findings indicate that MLM-based models might reach a diminishing return point as the supervised data size increases significantly. 
\end{abstract}

\section{Introduction}
Language modeling has emerged as an effective pretraining approach in wide variety of NLP models. Multiple techniques have been proposed, including bi-directional language modeling~\cite{peters2018deep}, masked language models~\cite{devlin2018bert}, and variants of denoising auto-encoder approaches~\cite{lewis2019bart,raffel2019exploring,joshi2019spanbert}. Today, it is rare to examine a leaderboard without finding the top spots occupied by some variant of a pretraining method.\footnote{https://super.gluebenchmark.com/leaderboard}

The future of NLP appears to be
paved by pre-training a ‘universal’ contextual representation
on wikipedia-like data at massive scale. Attempts along this path have pushed the frontier to up 10$\times$ to the size of wikipedia~\cite{raffel2019exploring}. However, the success of these experiments is mixed: although improvements have been observed, the downstream task is usually data-limited. There is evidence that large-scale pretraining does not always lead to state-of-the-art results \cite{raffel2019exploring}, especially on tasks such as machine translation, where abundance of training data, and the existence of strong augmentation methods such as back translation might have limited the benefit of pretraining. 

This paper examines the pretraining benefits of downstream tasks as the number of training samples increases. To answer this question, we focus on multi-class text classification since: (i) it is one of most important problems in NLP with applications spanning multiple domains. (ii) large sums of training data exists for many text classification tasks, or can be obtained relatively cheaply through crowd workers \cite{snow2008cheap}. We choose three sentiment classification datasets: Yelp review~\cite{yelp}, Amazon sports and electronics review~\cite{ni2019justifying}, ranging in size from 6 to 18 million examples. \footnote{These
datasets are the largest 
publicly available classifiaction datasets that we are
aware of.} 

We finetune a RoBERTa model~\cite{liu2019roberta} with increments of the downstream dataset, and evaluate the performance at each increment. For example, on the Yelp dataset whose size is 6 million, we train the models on subsets of the data with each subset size being in the sequence (60k, 600K, 1.8M, 3M .., 6M). For comparison, we also train a vanilla BiLSTM, and another BiLSTM which uses pretrained Roberta token embeddings. We observe that when both models are trained on 1\% of the data, the gap between BiLSTM and RoBERTa models is at its peak, but as the training dataset size increases, the BiLSTM model accuracy keeps on increasing whereas RoBERTa's accuracy remain mostly flat. As the dataset size increases, the accuracy gap shrinks to within 1\%.

Our study suggests that collecting data and training on the target tasks is a solution worth considering, especially in production environments where accuracy is not the only considered factor, rather inference latency is often just as crucial. We benchmarked the inference 
latency of the these models on both CPU and GPU for different batch sizes, and as expected, we observe at least 20$\times$ speedup for the BiLSTM compared to the RoBERTa.  This paper provides new experimental evidence and discussions for people to rethink the MLM pre-training paradigm in NLP,
at least for resource rich tasks.

\section{Related Works}
Scaling the number of training examples has long been identified as source of improvement for machine learning models in multiple domains including NLP~\cite{banko2001scaling}, computer vision~\cite{deng2009imagenet,sun2017revisiting} and speech ~\cite{amodei2016deep}.  Previous work has suggested that deep learning scaling may be predictable empirically \cite{hestness2017deep}, with model size  scaling sub-linearly with training data size. \cite{sun2017revisiting} concluded that accuracy increases logarithmally with respect to training data size. However, these studies have focused on training models in the the fully supervised setting, without pretraining. 

One closer work is \cite{he2019rethinking} where it is shown that randomly initialized standard computer-vision models perform no worse than their \textit{ImageNet} pretrained counterparts. However,
our work focuses on text classification. We do not examine the benefit of pretraining, at large, rather we focus on the benefit of pretraining for resource rich tasks. Another concurrent work that is still under review, in~\cite{DDD} observes that, in some translation task such as IWSLT’14, small language models exhibit even lower test loss compared to the large transformer model when the number of training samples increases.

\section{Experiments}

\subsection{Task and Data}

We focus on a multi-class sentiment classification task: given the user reviews, predict the rating in five points scale $\{1, 2, 3, 4, 5\}$. The experiments are conducted on the following three benchmark datasets.
\begin{itemize}
    \item \textbf{Yelp Challenge}~\cite{yelp} contains text reviews, tips, business and check-in sets in Yelp. We use the 6.7m user reviews with ratings as our dataset.
    \item \textbf{Amazon Reviews}~\cite{ni2019justifying} contains product reviews (ratings, text, helpfulness votes) from Amazon. We choose two categories: \textit{sports / outdoors}, and \textit{electronics} as two separate datasets. We only use the review text as input features.
\end{itemize}
The distribution across five ratings of each dataset is illustrated in Table \ref{data}. In our experiment, all the above data is split into 90\% for training and 10\% for testing.

\begin{table}[h]
\resizebox{0.48\textwidth}{!}{%
\begin{tabular}{c c c c c c c}
\hline
Dataset & Size & 1 & 2 & 3 & 4 & 5 \\
\hline
Yelp & 6.69M & 15\% & 8\% & 11\% & 22\% & 44\% \\
Sports & 11.9M & 7\% & 5\% & 7\% & 16\% & 65\% \\
Electronics & 18.6M & 11\% & 5\% & 7\% & 16\% & 61\% \\
\hline
\end{tabular}}
\caption{Data size and percentage of samples in each (n)-star category}
\label{data}
\end{table}

\begin{figure*}[t]
  \centering
  \includegraphics[width=6.2in]{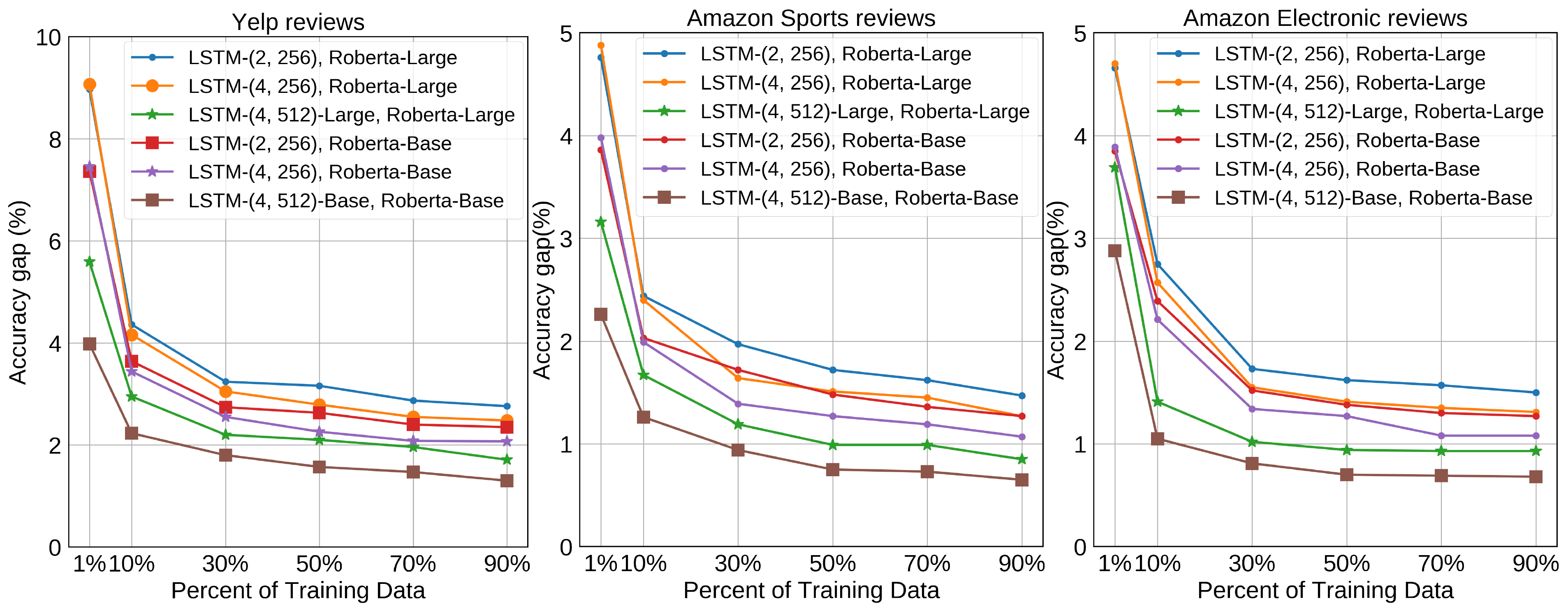}
  \caption{Accuracy Gap of Roberta, BiLSTM trained on different amount of data}
  \label{fig:size_impact}
\end{figure*}

\begin{table*}[h]
\centering
\resizebox{0.85\textwidth}{!}{%
\begin{tabular}{l|c c c c c c|c}
\hline
\multirow{2}{*}{Models} & \multicolumn{2}{c}{Yelp} & \multicolumn{2}{c}{Sports} & \multicolumn{2}{c|}{Electronics} & \multirow{2}{*}{Params}\\
 & Accuracy & $\Delta$ & Accuracy & $\Delta$ & Accuracy  & $\Delta$ & \\
\hline
Roberta-Large & 78.85 & - & 79.65 & - & 79.07 & - & 304M\\
Roberta-Base & 78.44 &  0.41  & 79.45 & 0.20 & 78.84 &   0.23 & 86M\\

LSTM-4-512 + Large  & 77.14 & 1.71 & 78.80 & 0.85 & 78.16 & 0.92 & 25M\\
LSTM-4-512 + Base  & 77.07 & 1.78 & 78.72 &  0.93 & 78.07 & 1.0& 24M\\

LSTM-4-256 + Large  & 77.02 & 1.83 & 78.76& 0.89 & 78.12 & 0.95 & 7.4M\\
LSTM-4-256 + Base   & 77.03 & 1.82 & 78.62 & 1.03 & 77.98 & 1.09& 6.8M\\
LSTM-4-256  & 76.37 & 2.48 & 78.38 & 1.27 & 77.76 & 1.31& 4.8M\\
LSTM-2-256  & 76.09 & 2.76  & 78.18 & 1.47 & 77.57 & 1.5 & 2.4M \\
\hline
\end{tabular}
}
\caption{Test Accuracy of Roberta-base, BiLSTM, and BiLSTM with Roberta Pretrained Token Embedding when
trained on the full dataset. The $\Delta$ column 
shows the difference between each model's accuracy and that of Roberta-Large. For LSTM models, 
LSTM-n-k denotes an LSTM model with n layers and k
cells. + Large or + Base indicate the use of 
Roberta Large or Roberta Base token embeddings, 
respectively. The number of parameters does not count the size of embedding table.}
\label{ac_results}
\end{table*}

\subsection{Models}
We choose the following three types of pretrained and vanilla models:
\begin{itemize}
    \item \textbf{RoBERTa} \cite{liu2019roberta} RoBERTa is a transformer-based model pretrained with masked language modeling objectives on a large corpus. We finetune our classification task on both Roberta-Base (12 layers, 768 hidden, 12 heads) and Roberta-Large (24 layers, 1024 hidden, 16 heads).
    \item \textbf{LSTM} \cite{hochreiter1997long} 
    We use a bidirectional LSTM with a max-pooling layer on top of the hidden states,
    followed by a linear layer. Token 
    embeddings of size 128 are randomly initialized.
    \item \textbf{LSTM + Pretrained Token Embedding} Similar to the previous setup,
    except we initialized the token embeddings with Roberta pretrained token embedding (Base: 768-dimensional embedding, Large: 1024-dimensional embedding). The embeddings are frozen during training.
\end{itemize}
For fair comparison, all the above models share the same vocabulary and BPE tokenizer \cite{sennrich2015neural}.

\subsection{Experimental Setup}
We use the Adam optimizer and the following hyperparameter sweep for each model. (i) RoBERTa is finetuned with the following learning rates $\{5e-6, 1e-5, 1.5e-5, 2e-5\}$, with linear warm up in the first 5\% of steps followed by a linear decay to 0. The batch size is set to 32, with dropout being 0.1. (ii) For the LSTM, it is trained with a constant learning rate from the sequence: $\{2.5e-4, 5e-4, 7.5e-4, 1e-3\}$. The batch size is set to 64. We train each model on 8 GPUs for 10 epochs and perform early stopping based on accuracy on the test set. The maximum sequence length of input was set to 512 for all models.

\section{Results}
\subsection{Impact of Data Size}

We first investigate the effect of varying the number of training samples, for fixed model and training procedure. We train different models using $\{1\%, 10\%, 30\%, 50\%, 70\%, 90\%\}$ amount of data to mimic the ``low-resource'', ``medium-resource'' and ``high-resource'' regime. Figure~\ref{fig:size_impact} shows that the accuracy delta between the LSTM and RoBERTa models at different percentages of the training data. From the plot, we observe the following phenomena: 

(i) Pretrained models exhibit a diminishing return behavior as 
the size of the target data grows. When we increase the number of training examples, the accuracy gap between Roberta and LSTM shrinks. For example,
when both models are trained with 1\% of the Yelp dataset, 
the accuracy gap is around 9\%. However, as we increases the amount of training data to 90\%, the accuracy gap drops to within 2\%. 
The same behaviour is observed on both Amazon review datasets, with 
the initial gap starting at almost 5\% for 1\% of the training data, then
shrinking all the way to within one point when most of the training data is used.


(ii) Using the pretrained RoBERTa token embeddings can further reduce the accuracy gap especially when training data is limited. For example, in the Yelp review data, a 4-layers LSTM with pretrained embeddings provides additional 3 percent gain compared to its counterparts. As Table \ref{ac_results} shows, an LSTM with pretrained RoBERTa token embeddings always outperforms the ones with random token initialization. This suggests 
that the embeddings learned during pretraining RoBERTa may constitute an efficient approach for transfer learning the knowledge learned in these large MLM. 

We further report the accuracy metric of each model using all the training data. 
The full results are listed in Table ~\ref{ac_results}. 
We observe that the accuracy gap is less than 1\% on the Amazon datasets.
even compared to 24 layers RoBERTa-large model. As for the Yelp dataset,
the accuracy gap is within 2 percent from the RoBERTa-large model, despite an 
order of magnitude difference in the number of parameters.

\subsection{Inference Time}

We also investigate the inference time of the three type of models on GPU and CPU. 
The CPU inference time is tested on Intel Xeon E5-2698 v4 with batch size 128. 
The GPU inference time is tested on NVIDIA Quadro P100 with batch size $\in\{128, 256, 384\}$. The maximum sequence length is 512. We run 30 times for each settings and take the average. The results are listed in TABLE~\ref{time_results}.

\begin{table}[htb]
\resizebox{0.47\textwidth}{!}{%
\begin{tabular}{l c c c c}
\hline
Model & CPU & \multicolumn{3}{c}{GPU} \\
\cline{3-5}
Batch size & 128 & 128 & 256 & 384\\
\hline
Roberta-Base & 323 & 16.1 & 16.1 & 16.1\\
Roberta-Large & 950 & 55.5 & 55.5 & - \\
\hline
LSTM-2-256 & 15.2 & 0.47 & 0.43 & 0.42\\
LSTM-4-256 & 28.1 & 1.17 & 0.94 & 0.86\\
LSTM-4-256+Base & 35.2 & 1.33 & 1.09 & 1.02\\
LSTM-4-256+Large & 37.5 & 1.33 & 1.17 & 1.07\\
LSTM-4-512+Base& 64.8 & 3.52 & 3.20 & 3.13\\
LSTM-4-512+Large& 64.8 & 3.36 & 3.32 & 3.26\\
\hline
\end{tabular}}
\caption{Inference time (ms) of Roberta, BiLSTM on CPU and GPU}
\label{time_results}
\end{table}

Not surprisingly, the LSTM model is at least 20 time faster even when compared to
the Roberta-Base. Note that the P100 will be out of memory when batch size is 384 for Roberta-Large. Another observation is that although using the Roberta pretrained token embedding 
introduces 10 times more model parameters compared to vanilla BiLSTM, the inference time only increases by less than 25\%. This is due to the most additional parameters are from a simple linear transformation.

\section{Discussion}
Our findings in this paper indicate that increasing the number of
training examples for `standard' models such as LSTM leads to performance gains that are within 1 percent of their massively pretrained counterparts. Due to the fact that there is no good large scale question answering dataset, it is not clear if the same findings would hold on this type of NLP tasks, which are more challenging and semantic-based. In the future work, we will run more experiments if there are some other large scale open datasets. Despite sentiment analysis being a crucial text classification task, it is possible, though unlikely, that the patterns observed here are limited to sentiment analysis tasks only. The rationale behinds that is that pretrained LSTMs have kept up very well with transformer-based counterparts on many tasks~\cite{radford2018improving}.

One way to interpret our results is that `simple' models have better regularization 
effect when trained on large amount of data, 
as also evidenced in the concurrent work~\cite{DDD}.The other side of the argument in interpreting our results is that MLM based pretraining still leads to 
improvements even as the data size scales into the millions. In fact, with a pretrained model and 2 million training examples, 
it is possible to outperform an LSTM model 
that is trained with 3$\times$ more examples.

\section{Conclusion}
Finetuning BERT-style models on resource-rich downstream tasks is not well studied. In this paper, we reported that, when the downstream task has sufficiently large amount of training exampes, i.e., millions, competitive accuracy results can be achieved by training a simple LSTM, at least
for text classification tasks.
We further discover that reusing the token embeddings learned during BERT pretraining 
in an LSTM model leads to significant improvements.
The findings of this work have significant implications on both the practical
aspect as well as the research on pretraining. For industrial applications where
there is a trade-off typically between accuracy and latency, our findings 
suggest it might be feasible to gain accuracy for faster models by collecting more 
training examples.

\bibliography{anthology,acl2020}
\bibliographystyle{acl_natbib}
\newpage
\appendix
\onecolumn
\section{Detailed Results}
We list the detailed results in TABLE~\ref{st_results}.
\begin{table}[h!]
\begin{tabular}{c c|c|c c c c c c}
\hline
Model & Parameters& Data Sets & 1\% & 10\% & 30\% & 50\% & 70\% & 90\% \\ \hline
\multirow{3}{*}{Roberta-Base}  & \multirow{3}{*}{86M} & yelp & 74.82 & 77.22 & 77.87 & 78.13 & 78.27 & 78.44 \\ 
&  & sports & 77.33 & 78.67 & 78.99 & 79.19 &79.35& 79.45     \\ &
& electronics & 76.68 & 77.74 & 78.40 & 78.62 & 78.72 & 78.84 \\ \hline
\multirow{3}{*}{Roberta-Large}  & \multirow{3}{*}{304M} & yelp & 76.43 & 77.94 & 78.37 & 78.66 & 78.74 & 78.85    \\ 
&  & sports & 78.23 & 79.08 & 79.24 & 79.43 & 79.61 & 79.65  \\ & 
& electronics & 77.49 & 78.10 & 78.61 & 78.86 & 78.99 & 79.07  \\ \hline
\multirow{3}{*}{BiLSTM-2}  & \multirow{3}{*}{2.4M} & yelp             & 67.46 & 73.58 & 75.13 & 75.50 & 75.87 & 76.09   \\ 
&  & sports     & 73.47 & 76.64 & 77.27 & 77.71 & 77.99 & 78.18     \\ & 
& electronics     & 72.83 & 75.35 & 76.88 & 77.24  & 77.42 & 77.57   \\ \hline
\multirow{3}{*}{BiLSTM-4}  & \multirow{3}{*}{4.8M} & yelp             & 67.36 & 73.78 & 75.32 & 75.87 & 76.19 & 76.37 \\ 
&  & sports & 73.35 & 76.68 & 77.60 & 77.92 & 78.16 & 78.38 \\ & 
& electronics  & 72.79 & 75.53 & 77.06 & 77.45 & 77.64 & 77.76 
\\ \hline
\multirow{2}{*}{BiLSTM-4-512}  & \multirow{3}{*}{25M} & yelp           & 70.84 & 74.99 & 76.07  & 76.56 &  76.80 & 77.14 \\ 
 &  & sports & 75.07 & 77.41 & 78.05  & 78.44 & 78.62 & 78.80 \\ + Large & 
 & electronics & 73.80 & 76.69 & 77.59 & 77.92 & 77.92 & 78.16  \\ \hline
\end{tabular}
\caption{Test Accuracy of Roberta, BiLSTM trained on different amount of data}
\label{st_results}
\end{table}
\end{document}